\documentclass[11pt,a4paper]{article}
\usepackage[hyperref]{naaclhlt2019}
\usepackage{times}
\usepackage{latexsym}
\usepackage{lipsum}
\usepackage{url}
\usepackage{CJKutf8}
\usepackage{latexsym}
\usepackage{graphicx}
\usepackage{url}
\usepackage{float}
\usepackage{amsmath}
\usepackage{longtable}
\usepackage{rotating}
\usepackage{multirow}
\usepackage{subfigure}
\usepackage{amssymb}
\usepackage{times}
\usepackage{xcolor}
\usepackage{soul}
\usepackage[utf8]{inputenc}
\usepackage{threeparttable}
\usepackage{subfigure}
\usepackage[T1]{fontenc}
\usepackage{upquote}
\usepackage{booktabs}
\usepackage{helvet}  
\usepackage{courier} 
\usepackage{algorithm}  
\usepackage{algorithmic}
\usepackage{setspace}

\aclfinalcopy 

\title{Improving Domain Adaptation Translation \\ with Domain Invariant and Specific Information}

\author{Shuhao Gu\textsuperscript{1,2} \quad Yang Feng\textsuperscript{1,2*}  \quad Qun Liu\textsuperscript{3} \\
  \textsuperscript{1}Key Laboratory of Intelligent Information Processing \\
Institute of Computing Technology, Chinese Academy of Sciences (ICT/CAS) \\
  \textsuperscript{2}University of Chinese Academy of Sciences \\
  \textsuperscript{3}Huawei Noah’s Ark Lab, Hong Kong, China\\
  {\tt \textsuperscript{1,2}\{gushuhao17g, fengyang\}@ict.ac.cn }\\ 
  {\tt \textsuperscript{3}qun.liu@huawei.com} 
  }
  
\date{}

\begin{document}

\maketitle
\newcommand\blfootnote[1]{%
\begingroup 
\renewcommand\thefootnote{}\footnote{#1}%
\addtocounter{footnote}{-1}%
\endgroup 
}
\begin{abstract}
 
  
  In domain adaptation for neural machine translation, translation performance can benefit from separating features into domain-specific features and common features. In this paper, we propose a method to explicitly model the two kinds of information in the encoder-decoder framework so as to exploit out-of-domain data in in-domain training. In our method, we maintain a private encoder and a private decoder for each domain which are used to model domain-specific information. In the meantime, we introduce a common encoder and a common decoder shared by all the domains which can only have domain-independent information flow through. Besides, we add a discriminator to the shared encoder and employ adversarial training for the whole model to reinforce the performance of information separation and machine translation simultaneously. Experiment results show that our method can outperform competitive baselines greatly on multiple data sets. 
  \blfootnote{*Corresponding Author}

\end{abstract}

\section{Introduction}

Neural machine translation (NMT) \cite{kalchbrenner2013recurrent,cho2014learning,sutskever2014sequence,bahdanau2014neural,gehring2017convolutional} has made great progress and drawn much attention recently. Most NMT models are based on the encoder-decoder architecture,
where all the sentence pairs share the same set of parameters for the encoder and decoder which makes NMT models have a tendency towards overfitting to frequent observations (e.g., words, word co-occurrences, translation patterns), but overlooking special cases that are not frequently observed. However, in practical applications, NMT models usually need to perform translation for some specific domain with only a small quantity of in-domain training data but a large amount of out-of-domain data.
Simply combining in-domain training data with out-of-domain data will lead to overfitting to the out-of-domain data. Therefore, some domain adaptation technique should be adopted to improve in-domain translation.

Fortunately, out-of-domain data still embodies common knowledge shared between domains. And incorporating the common knowledge from out-of-domain data can help in-domain translation. \citet{britz2017effective} have done this kind of attempts and managed to improve in-domain translation.
The common architecture of this method is to share a single encoder and decoder among all the domains and add a discriminator to the encoder to distinguish the domains of the input sentences. The training is based on adversarial learning between the discriminator and the translation , ensuring the encoder can learn common knowledge across domains that can help to generate target translation. ~\citet{zeng2018multi} extend this line of work by introducing a private encoder to learn some domain specific knowledge. They have proven that domain specific knowledge is a complement to domain invariant knowledge and indispensable for domain adaptation. Intuitively, besides the encoder, the knowledge inferred by the decoder can also be divided into domain specific and domain invariant and further improvement will be achieved by employing private decoders.
 
In this paper, in order to produce in-domain translation with not only common knowledge but in-domain knowledge, we employ a common encoder and decoder among all the domains and also a private encoder and decoder for each domain separately. The differences between our method and the above methods are in two points: first, we employ multiple private encoders rather where all the domains only have one private encoder; second, we also introduce multiple private decoders contrast to no private decoder. This architecture is based on the consideration that out-of-domain data is far more than in-domain data and only using one private encoder and/or decoder has the risk of overfitting. Under the framework of our method, the translation of each domain is predicted on the output of both the common decoder and its private decoder. In this way, the in-domain private decoder has direct influence to the generation of in-domain translation and the out-of-domain decoder is used to help train the common encoder and decoder better which can also help in-domain translation. We conducted experiments on English$\rightarrow$Chinese and English$\leftrightarrow$German domain adaptation tasks for machine translation under the framework of RNNSearch \cite{bahdanau2014neural} and Transformer \cite{vaswani2017attention} and get consistently significant improvements over several strong baselines.

  \section{Related Work}
    
  The task of domain adaptation for NMT is to translate a text in-domain for which only a small number of parallel sentences is available. The main idea of the work for domain adaptation is to introduce external information to help in-domain translation which may include in-domain monolingual data, meta information or out-of-domain parallel data.
 
 To exploit in-domain monolingual data, ~\citet{gulccehre2015using} train a RNNLM on the target side monolingual data first and then use it in decoding.
 ~\citet{domhan2017using} further extend this work by training the RNNLM part and translation part jointly.
 ~\citet{sennrich2015improving} propose to conduct back translation for the monolingual target data so as to generate the corresponding parallel data. 
  ~\citet{zhang2016exploiting} employs the self-learning algorithm to generate the synthetic large-scale parallel data for NMT training.
 To introduce meta information, ~\citet{chen2016guided} use the topic or category information of the input text to assistant the decoder and ~\citet{kobus2017domain} extend the generic NMT models, which are trained on a diverse set of data to, specific domains with the specialized terminology and style. 
 
 To make use of out-of-domain parallel data, \citet{luong2015stanford} first train an NMT model with a large amount of out-of-domain data, then fine tune the model with in-domain data.  
 \citet{wang2017sentence} select sentence pairs from the out-of-domain data set according to their similarity to the in-domain data and then add them to the in-domain training data. 
 \citet{chu2017empirical} construct the training data set for the NMT model by combining out-of-domain data with the over-sampled in-domain data.
 \citet{wang2017instance} combine the in-domain and out-of-domain data together as the training data but apply instance weighting to get a weight for each sentence pair in the out-of-domain data which is used in the parameter updating during back propagation.
\citet{britz2017effective} employ a common encoder to encode the sentences from both the in-domain and out-of-domain data and meanwhile add a discriminator to the encoder to make sure that only domain-invariant information is transferred to the decoder. They focus on the situation that the quantity of the out-of-domain data is almost the same as the in-domain data while our method can handle more generic situations 
and there is no specific demand for the ratio of the quantity between the in-domain and out-of-domain data. 
Besides, our method employs a private encoder-decoder for each domain which can hold the domain-specific features. In addition to the common encoder, \citet{zeng2018multi} further introduce a domain-specific encoder to each domain together with a domain-specific classifier to ensure the features extracted by the domain-specific encoder is proper. Compared to our method, they focus on the encoder and do not distinguish the information in the decoder.

  Adversarial Networks have achieved great success in some areas~\cite{ganin2016domain,goodfellow2014generative}. Inspired by these work, we also employ a domain discriminator to extract some domain invariant features which has already shown its effectiveness in some related NLP tasks. \citet{chen2017adversarial} use a classifier to exploit the shared information between different Chinese word segment criteria. \citet{gui2017part} tries to learn common features of the out-domain data and in-domain data through adversarial discriminator for the part-of-speech tagging problem. \citet{kim2017cross} train a cross-lingual model with language-adversarial training to generate the general information across different languages for the POS tagging problem. All these work try to utilize a discriminator to distinguish invariant features across the divergence.
  

\section{RNN-based NMT model}\label{sec:bgd}
\begin{figure}[!t]
    \centering
    \includegraphics[scale=0.3]{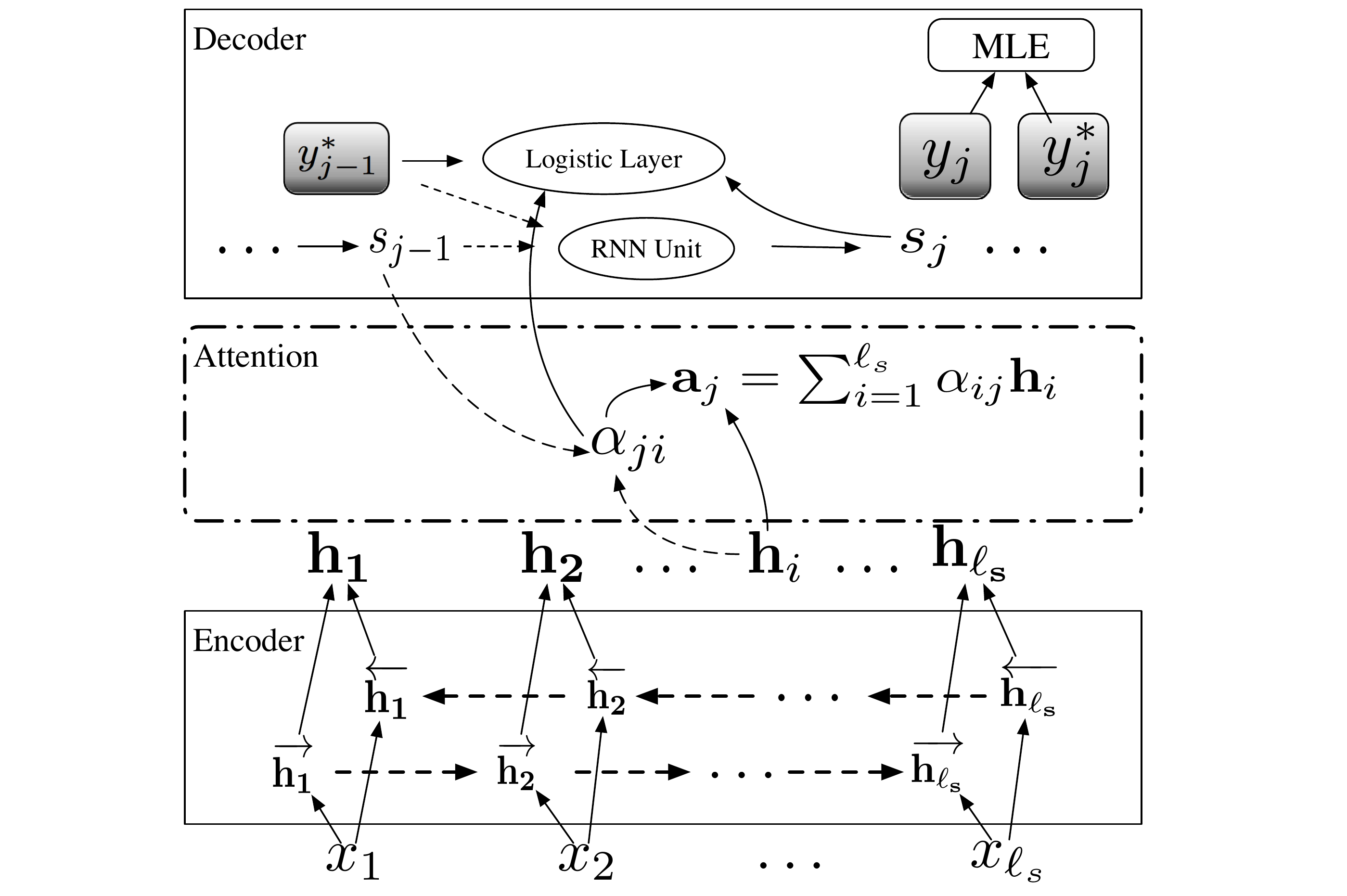}
    \caption{The architecture of the attention-based NMT.}
    \label{fig:nmt}
  \end{figure}

  Our method can be applied to both the RNN-based NMT model \cite{bahdanau2014neural} and self-attention-based NMT model \cite{vaswani2017attention}. In this paper, we will introduce our method under the RNN-based framework and the application to the self-attention-based framework can be implemented in a similar way. Before introducing our method, we will first briefly describe the RNN-based NMT model with attention shown in Figure~\ref{fig:nmt}.

  \textbf{The encoder} uses two GRUs to go through source words bidirectionally to get two hidden states\ $\overrightarrow {\bf h}_i$ and $\overleftarrow {\bf h}_i$ for the source word $x_i$, which are then concatenated to produce the final hidden states for $x_i$ as follows
  \begin{equation}
  {\bf h}_i = [\overrightarrow {\bf h}_i ; \overleftarrow {\bf h}_i]
  \end{equation}

  \textbf{The attention layer} aims to extract the source information which is most related to the generation of each target word. First it evaluates the correlation between the previous decoder hidden state ${\bf s}_{j-1}$\ and each source hidden state\  ${\bf h}_i$ by
  \begin{equation}\label{eq1}
    e_{ij} = {\bf v}_\alpha^T \tanh \left( {\mathbf W}_\alpha {\bf s}_{j-1} + {\mathbf U}_\alpha {\bf h}_i \right),
  \end{equation}
  next calculates $\alpha_{ij}$ which is the correlation degree to each target hidden state ${\mathbf h}_i$, and then gets the attention\ ${\mathbf c}_j$. The formulation is as follows  
  \begin{equation}\label{eq2}
    \alpha_{ij}=\frac{\exp(e_{ij})}{\sum_{i'=1}^{l_s}\exp({e_{i'j})}};\ \ \ \       
  {\mathbf c}_j = \sum_{i=1}^{l_s}\alpha_{ij}{\mathbf h}_i
  \end{equation}

  \textbf{The decoder} also employs a GRU  to get the hidden state $s_j$ for the target word $y_j$ as
  \begin{equation}\label{eq3}
    {\mathbf s}_j = g(y_{j-1}, {\mathbf s}_{j-1}, {\mathbf c}_j).
  \end{equation}
 Then the probability of the target word $y_j$ is defined as follows 
  \begin{equation}\label{eq4}
    p(y_j|{\mathbf s_j}, y_{j-1}, {\mathbf c_j}) \propto \exp(y_j^{\mathsf T}{\mathbf W}_o {\mathbf t}_j)
  \end{equation}
  where $t_j$ is computed by
  \begin{equation}\label{eq5}
    {\mathbf t}_j = {\mathbf U}_o{\mathbf s_{j-1}}+{\mathbf V}_o {\mathbf E} y_{j-1}+ {\mathbf C}_o{\bf c}_j
  \end{equation}

  \begin{figure}[!bt]
    \centering
    \includegraphics[scale=0.27]{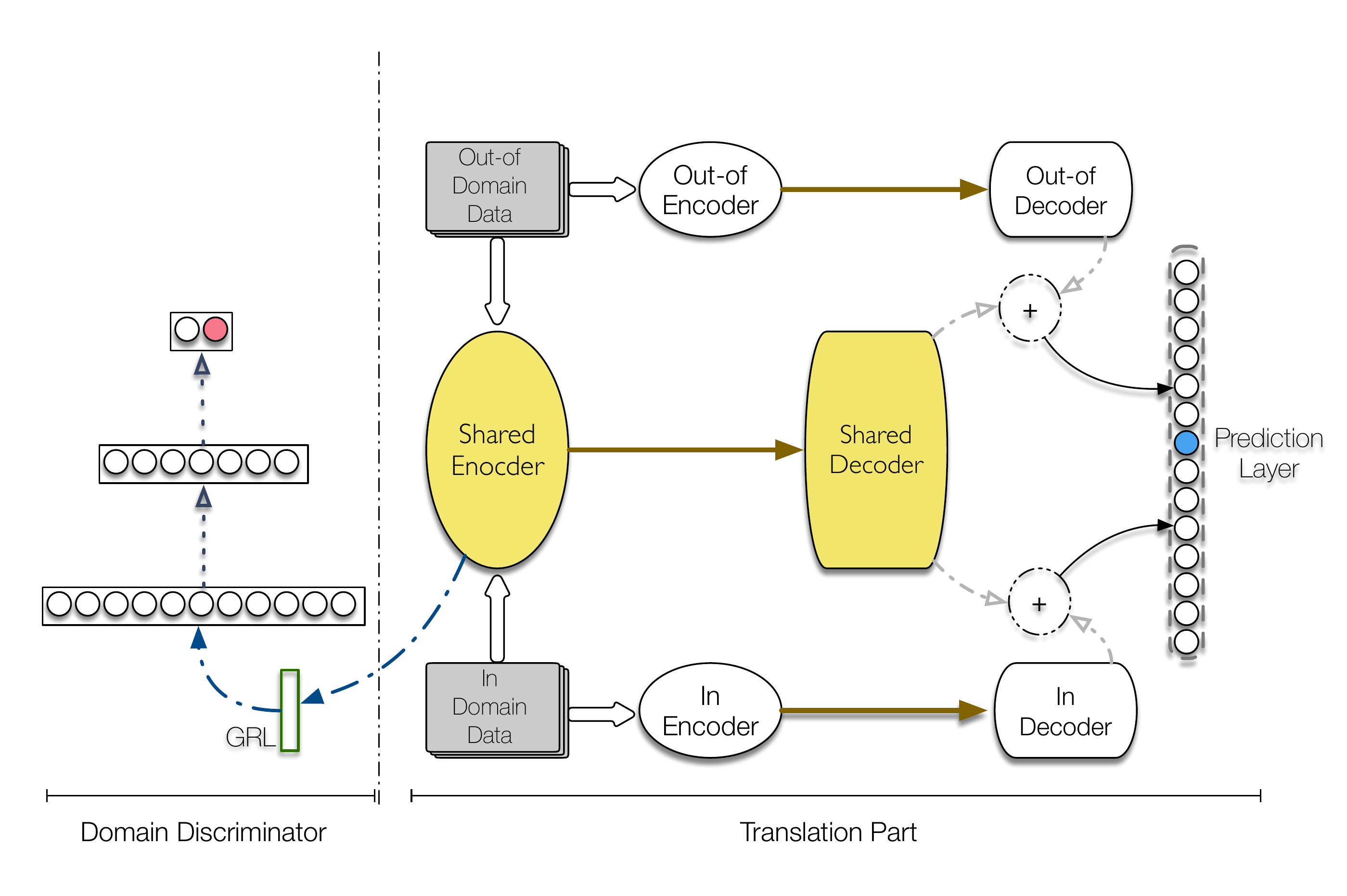}
    \caption{The architecture of the proposed method. GRL means the gradient reversal layer which will multiply a negative constant to the gradients during back-propagation.}
    \label{fig:our}
  \end{figure}

  \section{The Proposed Method}
  
  Assume that we have two kinds of training data: out-of-domain and in-domain, and we want to get the translation for the in-domain input. The out-of-domain and in-domain data can be represented as
   \begin{equation}\label{eq7}
    \begin{split}
    &{\bf out}  = { \{({\bf x}^k, {\bf y}^{*k})\}}_{k=1}^{N_{\bf out}} \sim {\mathcal D_{\bf out} };\\
    &{\bf in}  = { \{({\bf x}^k, {\bf y}^{*k})\}}_{k=1}^{N_{\bf in}} \sim {\mathcal D_{\bf in} }
    \end{split}
  \end{equation}
  
  The main idea of our method is to extract domain invariant information from the out-of-domain data to improve in-domain translation. To this end, we employ a common encoder and a common decoder shared by both of the domains, and a private encoder and a private decoder for each domain. The main architecture given in Figure~\ref{fig:our}. 
  
  The working scenario of our method is as follows. When a sentence comes, it is inputted into the shared encoder and the private encoder of the corresponding domain simultaneously. Then the output of the shared encoder is fed into the shared decoder and the output of the private encoder into its corresponding private decoder. Finally, the shared decoder and the private decoder collaborate together to generate the current target word with a gate to decide the contribution ratio. 
  
  In addition, our method also introduce a discriminator to distinguish the domain of the input sentence based on the output of the shared encoder. When the discriminator cannot predict the domain of the input sentence, we can think the knowledge encoded in the shared encoder is domain invariant. This is achieved with a gradient reversal layer (GRL) so that the gradients are reversed during back-propagation. In this way, the adversarial training is performed between the translation and the discriminator. 
  
  
  \subsection{The Translation Part}
 
 {\bf The Encoder}
 
  Our model has a shared encoder, an in-domain private encoder and an out-of-domain private encoder, where the shared encoder accepts input from the two domains. Given a sentence of domain $\mathrm{p}$ ($\mathrm{p} \in {\bf \{in, out\}}$), the shared encoder and the private encoder of domain $\mathrm{p}$ will roll the sentence as the encoder shown in Section~\ref{sec:bgd} and the outputs of the shared encoder and the private encoder for word $x_j$ are represented as ${\bf h}_j^\mathrm{c}$ and ${\bf h}_j^\mathrm{p}$ respectively.

  \noindent {\bf The Attention Layer}
  
  As the output of the shared encoder is only fed to the shared decoder and the output of the private encoder of domain $\mathrm{p}$ only flows to the private decoder of domain $\mathrm{p}$, we only need to calculate the attention of the shared decoder over the shared encoder and the attention of the private decoder of domain $\mathrm{p}$ over the private encoder of domain $\mathrm{p}$. We calculate these two attentions as in Section~\ref{sec:bgd} and denote them as ${\mathbf c}_j^\mathrm{c}$ and ${\mathbf c}_j^\mathrm{p}$ for the shared decoder and the private decoder, respectively.

  \noindent {\bf The Decoder}
  
  We also maintain a shared decoder, an in-domain private decoder and an out-of-domain private decoder
  For a sentence of domain $\mathrm{p}$ ($\mathrm{p} \in {\bf \{in, out\}}$), the shared decoder and the private decoder of domain $\mathrm{p}$ act in the same way as shown in Equation~\ref{eq3} and Equation~\ref{eq5} and then produce the hidden states  ${\mathbf s}_j^\mathrm{c}$ and ${\mathbf t}_j^\mathrm{c}$ for the shared decoder, and
  ${\mathbf s}_j^\mathrm{p}$ and ${\mathbf t}_j^\mathrm{p}$ for the private decoder.
  
  To predict the target word $y_j$, ${\mathbf t}_j^\mathrm{c}$ and ${\mathbf t}_j^\mathrm{p}$ are weighted added to get ${\mathbf t_j}$ as
    \begin{equation}
  \begin{split}
      &z_j = \sigma({\mathbf W}_z {\mathbf t}_j^{\mathrm{c}} + {\mathbf U}_z {\mathbf t}_j^{\mathrm p}); \\
      &{\mathbf t}_j = z_j \cdot {\mathbf t}_j^{\mathrm{c}}+ (1 - z_j) \cdot {\mathbf t}_j^{\mathrm p}
    \end{split}
  \end{equation}
  Where $\sigma(\cdot)$ is the sigmoid function and ${\mathbf W}_z$ and ${\mathbf U}_z$ are shared by in-domain and out-of-domain.
  Finally the probability of the target word $y_j$ is computed with
  \begin{equation}\label{eq10}
    P(y_j|\dots) \propto \exp(y_j^{\mathsf T}{\mathbf W}_o {\mathbf t}_j);
  \end{equation}

  \subsection{The Domain Discriminator}
  The domain discriminator acts as a classifier to determine the knowledge encoded in the shared encoder is from in-domain or from out-of-domain. When a well trained discriminator can't classify the domain properly, we can think the knowledge in the shared encoder is domain invariant~\cite{ganin2016domain}. As CNN has shown its effectiveness in some related classification tasks~\cite{zhang2015character,yu2017seqgan}, we construct our discriminator with CNN.
  
  First, the input to the CNN is the representation of the whole source sentence which is got by concatenating the sequence of hidden states generated by the shared encoder as
  \begin{equation}
    \Pi_{1:I} = \mathbf{h}_1 \oplus \mathbf{h}_2 \oplus \dots \oplus \mathbf{h}_I
  \end{equation}
  where $I$ is the length of the source sentence and $\mathbf{h}_1,...,\mathbf{h}_I$ is the hidden state of the corresponding source word. $\oplus$ stands for the concatenation operation of the hidden states, and we can get the final source sentence representation $\Pi_{1:I} \in \mathbb{R}^{I \times m}$ where $m$ is the dimension of the hidden state.
  
  We then employ a kernel ${\mathnormal w} \in \mathbb{R}^{l \times m}$ to apply a convolutional operation to produce a new feature map:
  \begin{equation}\label{eq:feature}
    {\mathbf f} = \rho ({\bf {\mathnormal w}} \otimes \Pi_{1:I} + b_f)
\end{equation}
where $\rho$ is the ReLU activation function, $\otimes$ stands for the convolutional operation of the kernel and b is the bias term. A number of different kinds of kernels with different windows sizes are used in our work to extract different features at different scales. Next, we apply a max-over-time pooling operation over the feature maps to get a new feature map.
To further improve the performance of the discriminator, following the work~\cite{yu2017seqgan}, we also add the highway architecture~\cite{srivastava2015highway,zhang2018refining} behind the pooled feature maps where we use a gate to control the information flow between the two layers.
Finally, the combined feature map is fed into a fully connected network with a sigmoid activation function to make the final predictions:
\begin{equation}
    p(d) \propto \exp({\mathbf W}_d \cdot {\mathbf f} + b_d);
\end{equation}
where $d$ is the domain label of in-domain or out-of-domain.

\section{Training}\label{loss}
Our final loss considers the translation loss and the domain prediction loss. For the translation loss, we employ cross entropy to maximize the translation probability of the ground truth, so we have this loss as follows and the training objective is to minimize the loss.

  \begin{equation}
    \mathcal L_{\bf MT} = -\sum_{k=1}^{N_{\bf in}  + N_{\bf out} } \sum_{j=1}^{J^k} \log p(y_j^{*k}) 
  \end{equation}
  where $N_{\bf in} $ and $N_{\bf out} $ are the number of training sentences for in-domain and out-of-domain data respectively, $J^k$ is the length of the $k$-th ground truth sentence, and $p(y_j^{*k})$ is the predicted probability of the $j$-th word for the $k$-th ground truth sentence. 
  
  Note that we have three different encoders and three different decoders in total, including the shared encoder and decoder, the in-domain private encoder and decoder, and the out-of-domain private encoder and decoder, and all of them have their own parameters.
  
  For the domain prediction loss, we also use cross-entropy to minimize the following loss
    \begin{equation}\label{eq12}
    \mathcal L_{\mathfrak D} = -\sum_{k=1}^{N_{\bf in}  + N_{\bf out}} \log p(d^{*k})
  \end{equation}
  where $d^{*k}$ is the ground truth domain label of the $k$-th input sequence.
  
Then the final loss is defined as
  \begin{equation}\label{eq13}
  \mathcal L= \mathcal L_{\bf MT} +  \lambda  \mathcal L_{\mathfrak D} 
  \end{equation}
  where $\lambda$ is a hyper-parameter to balance the effects of the two parts of loss. We gradually tried $\lambda$ from 0.1 to 2.5 and set it to 1.5 in our final experiments.
  
  Borrowing ideas from \citet{ganin2016domain}, we introduce a special gradient reversal layer (GRL) between the shared encoder and the domain discriminator. During forward propagation, the GRL has no influence to the model, while during back-propagation training, it multiplies a certain negative constant to the gradients back propagated from the discriminator to the shared encoder. In this way, an adversarial learning is applied between the translation part and the discriminator.
  
  At the beginning of the training, we just use the $\mathcal L_{\bf MT}$ to train the translation part on the combined data, including the shared encoder-decoder and the in-domain and out-of-domain private encoder-decoder. Then we use $\mathcal L_{\mathfrak D}$ to only train the domain discriminator until the precision of the discriminator reach 90\% while the parameters of the shared encoder keep unupdated.
  Finally, we train the whole model with the complete loss $\mathcal L$ with all the parameters updated. In the training process, the sentences in each batch is sampled from in-domain and out-of-domain data at the same rate.


  During testing, we just use the shared encoder-decoder and the private in-domain encoder-decoder to perform in-domain translation.

  \section{Experiments}
    We evaluated our method on the English$\rightarrow$Chinese (En-Zh),  German$\rightarrow$English (De-En) and English$\rightarrow$German (En-De) domain adaptation translation task.


   \subsection{Data Preparation}
    \textbf{English$\rightarrow$Chinese} 
    For this task, out-of-domain data is from the LDC corpus\footnote{https://www.ldc.upenn.edu/} that contains 1.25M sentence pairs. The LDC data is mainly related to the \textbf {News} domain. We chose the parallel sentences with the domain label \textbf {Laws} from the UM-Corpus~\cite{tian2014corpus} as our in-domain data. We chose 109K, 1K and 1K sentences from the UM-Corpus randomly as our training, development and test data. We tokenized and lowercased the English sentences with Moses\footnote{http://www.statmt.org/moses/} scripts. For the Chinese data, we performed word segmentation using Stanford Segmenter\footnote{https://nlp.stanford.edu/}. 
    
\textbf{German$\rightarrow$English }
  For this task, the training data is from the Europarl corpus distributed for the shared domain adaptation task of WMT 2007~\cite{callison2007meta} where the out-of-domain data is mainly related to the \textbf {News} domain, containing about 1.25M sentence pairs, and in-domain data is mainly related to the \textbf {News Commentary} domain which is more informal compared to the news corpus, containing about 59.1K sentences.
  We also used the development set of the domain adaptation shared task. Finally, we tested our method on the NC test set of WMT 2006 and WMT 2007. We tokenized and lowercased the corpora.
    
\textbf{English$\rightarrow$German}
  For this task, out-of-domain corpus is from the WMT 2015 en-de translation task which are mainly \textbf {News} texts~\cite{Bojar2015FindingsOT} containing about 4.2M sentence pairs. For the in-domain corpus, we used the parallel training data from the IWSLT 2015 which is mainly from the the \textbf {TED talks} containing about 190K sentences. In addition, dev2012 and test2013/2014/2015 of  IWSLT 2015 were selected as the development and test data, respectively. We tokenized and truecased the corpora. 

Besides, 16K, 16K and 32K merging operations were performed to learn byte-pair encoding(BPE)~\cite{sennrich2015neural} on both sides of the parallel training data and sentences longer than 50, 50 and 80 tokens were removed from the training data, respectively.

   \subsection{Systems} 
   We implemented the baseline and our model 
  by PyTorch framework\footnote{http://pytorch.org}. 
  For the En-Zh and De-En translation task, batch size was set to 80 and vocabulary size was set to 25k which covers all the words in the training set. The source and target embedding sizes were both set to 256 and the size of the hidden units in the shared encoder-decoder RNNs was also set to 256. During experiments, we found that the shared encoder-decoder played a major role in the model and the size of the private encoder-decoder didn’t influence the results too much.  Thus we just set the size of the  private encoder-decoder one-quarter of the shared encoder-decoder considering the training and decoding speed.
  
  For the En-De translation task, batch size was set to 40 and vocabulary size was set to 35K in the experiment. The source and target embedding sizes were both set to 620 and the size of the hidden units in the shared encoder and decoder RNNs was set to 1000. As mentioned before, the size of the private encoder-decoder was just one-quarter of the shared encoder-decoder.
  
  
  All the parameters were initialized by using uniform distribution over $[-0.1, 0.1]$. The adadelta algorithm was employed to train the model. We reshuffled the training set between epochs. Besides, the beam size was set to 10.
  
  {\textbf{Contrast Methods}} We compared our model with the following models, namely:
  
  \textbullet \ \textbf{In} : This model was trained only with the in-domain data.
  
  \textbullet \ \textbf{Out + In} : This model was trained with both of the in-domain and out-of-domain data.
  
  \textbullet \ \textbf{Sampler}~\cite{chu2017empirical} : This method over-sampled the in-domain data and concatenated it with the out-of-domain data. 
  
  \textbullet \ \textbf{Fine Tune}~\cite{luong2015stanford} :  This model was trained first on the out-of-domain data and then fine-tuned using the in-domain data.
  
  \textbullet \ \textbf{Domain Control (DC)}~\cite{kobus2017domain} : This method extend word embedding with an arbitrary number of cells to encode domain information.
  
  \textbullet \ \textbf{Discriminative Mixing (DM)}~\cite{britz2017effective} : This method adds a discriminator on top of the encoder which is trained to predict the correct class label of the input data. The discriminator is optimized jointly with the translation part.
  
  \textbullet \ \textbf{Target Token Mixing (TTM)}~\cite{britz2017effective} : This method append a domain token to the target sequence. 
  
  \textbullet \ \textbf{Adversarial Discriminative Mixing(ADM)}~\cite{britz2017effective} : This method is similar with our model which also add a discriminator to extract common features across domains. 
  The biggest difference is that we add private parts to preserve the domain specific features. Besides we also applied a different training strategy as the section~\ref{loss} describes so that our method can handle more generic situations.
  
  Noting that our model has a private encoder-decoder which brings extra parameters, we just slightly extend the hidden size of the contrast model to make sure that the total parameter number of the contrast model is equal to the number of our model's translation part. 

  \begin{table}[h!]
    \centering
    \renewcommand\arraystretch{1.3}
    \begin{tabular}{l|c|c|c}
    \hline
    \hline
     \textbf {En-Zh} & dev & test & average\\
    \hline
    In  &32.45 & 30.42 & 31.44\\
    Out + In  &30.37 & 28.76 & 29.57\\
    Sampler & 35.06 & 32.97 & 34.02 \\
    Fine Tune \ \ \ \ & 35.02 & 33.36 & 34.19 \\
    DC & 31.08&   29.59& 30.34\\
    DM & 30.98& 29.73 & 30.36 \\
    TTM & 31.77 & 30.11 & 30.94 \\
    ADM & 31.23 & 29.88 & 30.56\\
    \hline
    our method & 36.55**& 34.84**& 35.70\\
    \hline
    \end{tabular}
    \caption{Results of the en-zh translation experiments. The marks indicate whether the proposed methods were significantly better than the best performed contrast models(**: better at significance level$ \alpha$=0.01, *:$\alpha$=0.05)\cite{collins2005clause}}
    \label{table2}
  \end{table}

\begin{figure*}[bt]
  \centering
  \subfigure[without discriminator]{
  \includegraphics[scale=0.335]{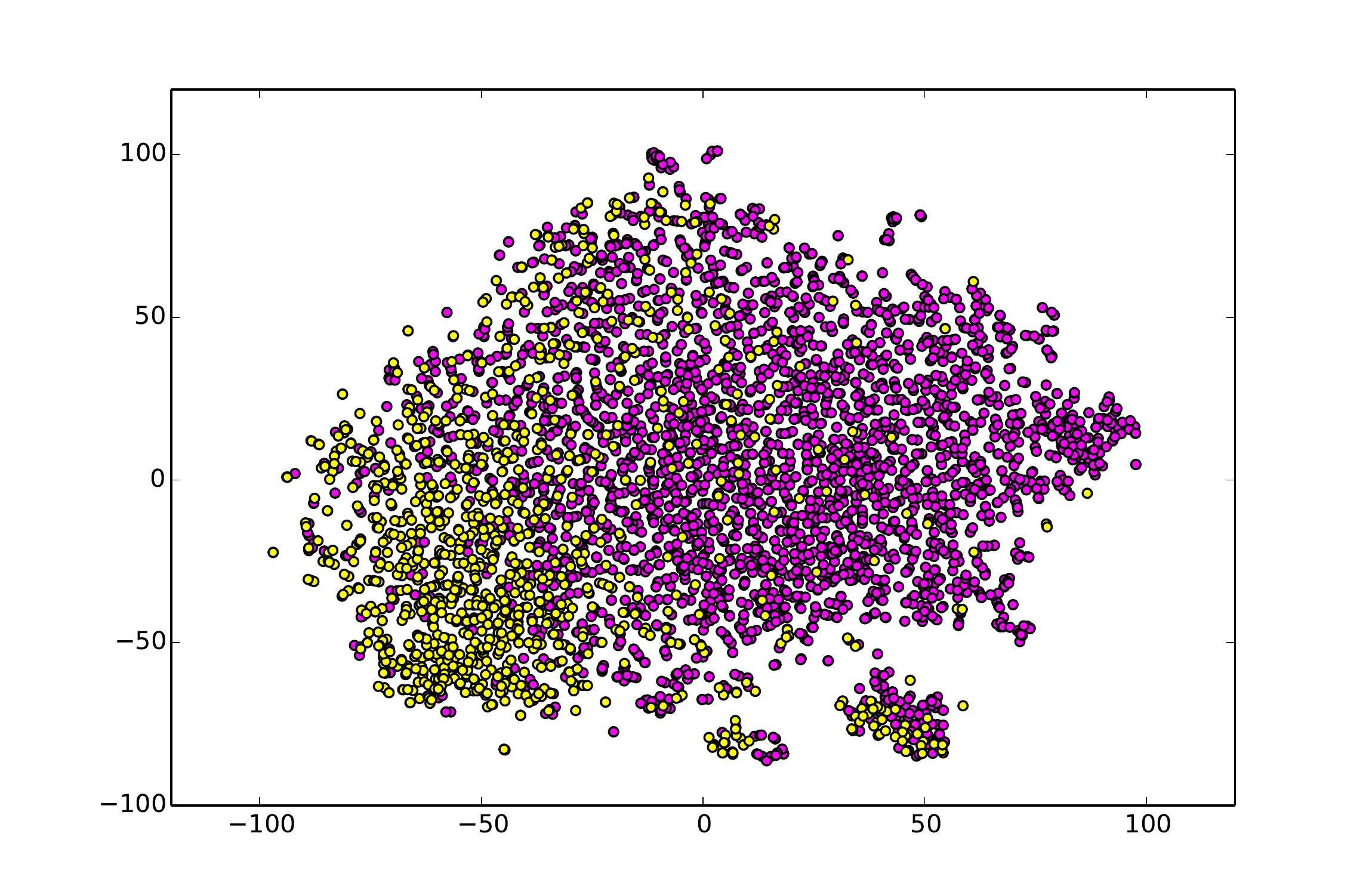}
  }
  \subfigure[full model]{
  \includegraphics[scale=0.335]{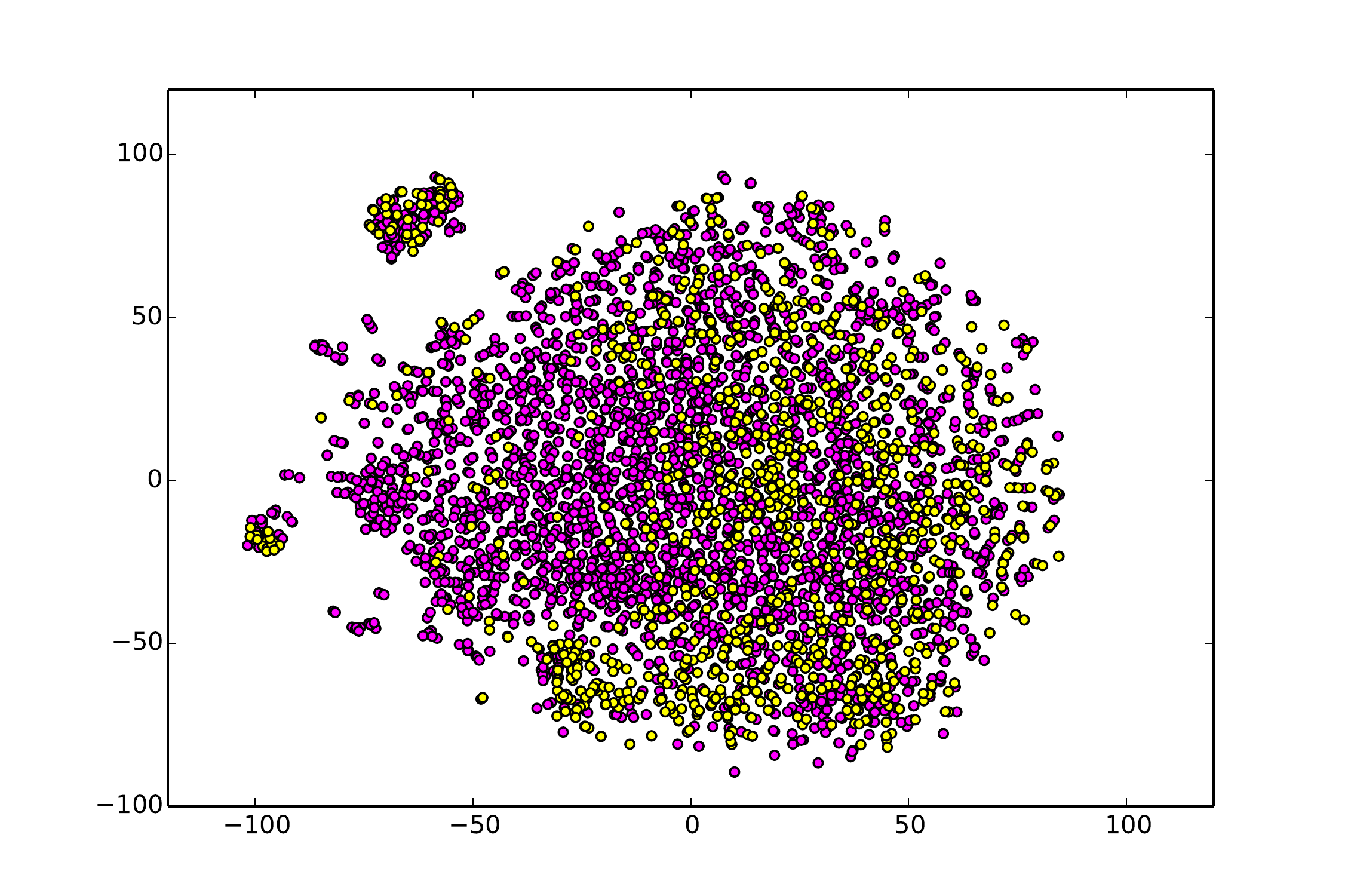}
  }
  \caption{The shared encoder's hidden state of the two models. Data from the out-of-domain are presented as Pink dots while data from the in-domain are presented as Yellow dots. There is an obvious separation of the results of the without discriminator model but the hidden states of the shared encoder of our full model are well-distributed.}
  \label{fig:scatter}
  \end{figure*}

  \subsection{Main Results}

    The \textbf{En-Zh Experiments} Results are measured using char based 5-gram BLEU score~\cite{papineni2002bleu} by the {\em multi-bleu.pl} script. The main results are shown in Table~\ref{table2}. On both of the development set and test set, our model significantly outperforms the baseline models and other contrast models. Furthermore, we got the following conclusions:
   
   First, the baseline model 'In' surpass the 'Out + In' model which shows that the NMT model tends to fit out-of-domain features if we directly include out-of-domain data into the in-domain data, as the domain specific features embodied in out-of-domain data is in greater quantity than that in in-domain data. However, we also found that the model will over fit so soon if we only use the in-domain data so it is necessary to make use of the out-of-domain data to improve the translation performance.
   
   \begin{table}[h!]
    \centering
    \renewcommand\arraystretch{1.3}
    \begin{tabular}{l|c|c|c}
    \hline
    \hline
     \textbf { De-En} & test06 & test07 & average\\
    \hline
    In  &23.36 & 25.00 & 24.18\\
    Out + In  &20.69 & 22.43 & 21.56\\
    Sampler & 26.83 & 29.01 & 27.92 \\
    Fine Tune \ \ \ \ & 27.02 & 29.19 & 28.11 \\
    \hline
    our method & 27.97*& 30.67**& 29.32\\
    \hline
    \end{tabular}
    \caption{Results of the WMT 07 De-En translation experiments.}
    \label{tablewmt7}
  \end{table}
   Second, we found that when the in-domain data is much less than the out-of-domain data, some contrast methods for domain adaptation, such as DC, DM TTM and ADM, didn't perform well. They were worse than the baseline model 'in' and only slightly better than 'out + in'.
   These methods all try to take domain information into translation in their own ways which actually brings improvement compared with the 'out + in' model. However, as the out-of-domain data is much more than the in-domain data, the model will still tends to fit out-of-domain data and ignore the in-domain information which will degrade the final performance. Therefore, it is necessary to handle the in-domain data separately in some way.
   The 'Sampler' and 'Fine Tune' perform better because they receive much more information from the in-domain data compared with other methods, but they don't make use of the domain information when translating.

   Last, our model achieves the best performance among all the contrast models. The shared encoder extract the domain invariant features of the two domains with the help of the discriminator so that the shared part will be well trained using all the in-domain and out-of-domain data. 
   At the meantime, we also consider the domain specific features and the private encoder-decoder can receive enough information from the in-domain data to prevent the whole model from overfitting the out-of-domain features. 
   
   \begin{table}[tbp]
    \centering
    \renewcommand\arraystretch{1.3}
    \begin{tabular}{l|c|c|c}
    \hline
    \hline
    {\bf  En-De} & test13 & test14  & test15 \\
    \hline
    In  &25.83  &21.97   &24.64 \\
    Out + In  &26.45 & 23.21 & 25.85\\
    Sampler &29.70  &25.71  &28.29 \\
    Fine Tune & 30.48 & 26.55  & 28.62 \\
    \hline
    \citet{sennrich2015improving} & 28.20 & 24.40 & 26.70 \\
    \citet{wang2017instance} & 28.58 & 24.12 & -  \\
    \hline
    our method &30.99 & 26.94  & 29.30* \\
    \hline
    \end{tabular}
    \caption{Results of the IWSLT 15 En-De experiments. The second part results were directly taken from their papers.}
    \label{table3}
  \end{table}

    The \textbf { De-En Experiments}  and \textbf { En-De Experiments} results are shown in the Table~\ref{tablewmt7} and~Table~\ref{table3}. Results are measured using word based 4-gram BLEU score~\cite{papineni2002bleu} by the {\em multi-bleu.pl} script. In these two experiments, we only compared our method with the baseline model and the competitive contrast methods 'Sampler'~\cite{kobus2017domain} and 'Fine Tune'~\cite{luong2015stanford}.
   Similar to the previous experiment results, our method still achieves the best performance compared to all contrast models, which demonstrates again that our model is effective and general to different language pairs and different domains.

\subsection{Experiment Analysis}
We made some some detailed analysis to empirically show the effectiveness of our model based on En-Zh translation task.
  \subsubsection{Ablation Study}
  In order to further understand the impact of the components of the proposed model, we performed some further studies by training multiple versions of our model by removing some components: The first model removed the domain discriminator but preserved the private part. The second one removed the private encoder-decoder but kept the domain discriminator.  The last one just removed both of those two parts.


   Results are shown in the Table~\ref{table5}. As expected, the best performance is obtained with the simultaneous use of all the tested elements. When we removed the private encoder-decoder, the result shows that the score was reduced by 0.79, which indicates that our private part can preserve some useful domain specific information which is abandoned by the shared encoder.
   When we removed the discriminator, the result was reduced by 0.76. This result supports our idea that modeling common features from out-of-domain data can benefit in-domain translation.
    When we removed both of the two components, we got the lowest score. The total result shows that every component of our model plays an important role in our model.

   \begin{table}[tbp]
    \centering
    \renewcommand\arraystretch{1.4}
    \begin{tabular}{c c|c|c|c}
    \hline
    \hline
     DCN & Private & dev & test & average\\
    \hline
    $\surd$ & $\surd$ & 36.55& 34.84& 35.70 \\
    $\surd$ & $\times$  & 35.73& 34.09 & 34.91\\
    $\times$ & $\surd$ & 35.67& 34.22 & 34.94\\
    $\times$ & $\times$ &35.13 &33.36 & 34.25 \\
    \hline
    \end{tabular}
    \caption{Results of the ablation study. "DCN" means the discriminator and "Private" means the private encoder-decoder.}
    \label{table5}
  \end{table}

  \subsubsection{Impact of the Discriminator}
  To verify whether the discriminator have learned the domain invariant knowledge, we did the following experiments using model \textbf {without discriminator} and our \textbf{full model} with the domain discriminator of the former subsection.

  We sampled 3000 sentences randomly from the out-of-domain and 1000 sentences from the in-domain En-Zh parallel sentences as the test data. Then they were fed into the shared encoders of the two models to get the reshaped feature maps as the Equation~\ref{eq:feature} describes. Next, we used the t-Distributed Stochastic Neighbor Embedding(t-SNE)\footnote{http://lvdmaaten.github.io/tsne/} technique to do a dimensionality reduction to the hidden state. Results are shown in the Figure~\ref{fig:scatter}. 
  We also calculate the average value of the coordinates of each domain's hidden state. The results are shown in Table~\ref{table4}
  \begin{table}[!tbp]
  \centering
  {\footnotesize
  \renewcommand\arraystretch{1.2}
  \begin{tabular}{|l||c|c|c|}
  \hline
  {} & Out-of-Domain & In-Domain & distance \\
  \hline
  - DCN & (12.9, 5.8) & (-38.9,-17.5) & 56.8 \\
  \hline
  Full Model & (-4.4, 3.0) & (13.2,-9.0) & 21.3 \\
  \hline
  \end{tabular}
  }
  \caption{The average coordinates value and its distance of the the hidden states. '- DCN' is the model without domain discriminator. }\label{table4}
  \end{table}
  
  \begin{table}[tbp]
    \centering
    \renewcommand\arraystretch{1.3}
    \begin{tabular}{l|c|c|c}
    \hline
    \hline
    {\bf  En-Zh} & test1 & test2  & test3 \\
    \hline
    Out + In  &22.31 & 18.82 & 17.59\\
    Sampler &21.60  &18.64  &16.93 \\
    Fine Tune & 13.18 & 11.94  & 11.55 \\
    \hline
    our method &22.61 & 19.36  & 17.78 \\
    \hline
    \end{tabular}
    \caption{Results of the out-of-domain translation task. The test sets are from the NIST test sets but we exchange the translation directions. }
    \label{table:out}
  \end{table}

  From the figure, we can find that here is an obvious separation in the results of the model without discriminator and the numerical analysis also support this point, which indicates that the shared encoder without the help of discriminator will treat the data from different domains differently. All the domain shared features and domain specific features are just mixed together. On the contrary, the output of the shared encoder of our full model is well distributed. This proves that the discriminator can help the shared encoder to extract domain invariant features 
  which then help to improve the translation performance in in-domain.

\subsubsection{Out-of-domain Translation Performance}
Despite the fact that the purpose of our work is to improve the in-domain translation performance, the domain invariant features extracted from the training data are also beneficial to the out-of-domain translation performance. To prove this, we use the NIST 03 04 and 05 test sets which are mainly related to the \textbf{News} domain as our out-of-domain test set. Noting that the origin set was designed for the Zh-En translation task and each sentence has four English references, we just chose the first reference as the source side sentence for our En-Zh translation task.
The results are shown in the Table~\ref{table:out}
We can conclude from the results that the "Fine Tune" method suffered a catastrophic forgetting caused by parameter shift during the training process. On the contrary, our method can achieve a mild improvement on the out-of-domain compared to the baseline system.


\subsubsection{Combined With Transformer Model}
Transformer~\cite{vaswani2017attention} is an efficient NMT architecture. To test the generality of our method, we also conducted relevant experiments based on the transformer model. 
We did the experiment based on the Fairseq code\footnote{https://fairseq.readthedocs.io/en/latest/index.html}. 
The implementation on this translation framework is similar with the way on the RNN based models.
The encoder and decoder of our final model consist 3 sublayers. The number of the multi-head attention was set to 4 and the embedding dim was set to 256. We also compared with the 'Sampler' and 'Fine Tune' method based on transformer. The results are shown in~\ref{tabletrans}. According to the table, our method still outperforms than other models, which can prove that our method has a good generality across different translation architecture.

\begin{table}[tbp]
    \centering
    \renewcommand\arraystretch{1.3}
    \begin{tabular}{l|c|c|c}
    \hline
    \hline
    {\bf En-Zh} & dev & test & average \\
    \hline
    In  &32.61  & 30.33  & 31.47\\
    Sampler &35.84  & 33.68 & 34.76 \\
    Fine Tune & 36.01 & 34.03  & 35.02 \\
    \hline
    our method & 37.26**& 35.39**  & 36.33 \\
    \hline
    \end{tabular}
    \caption{Results of the En-Zh experiments based on the transformer model.}
    \label{tabletrans}
  \end{table}

  \section{Conclusions}
  
  In this paper, we present a method to make use of out-of-domain data to help in-domain translation. The key idea is to divide the knowledge into domain invariant and domain specific. The realization way is to employ a shared encoder-decoder to process domain invariant knowledge and a private encoder-decoder for each domain to process knowledge of the corresponding domain. In addition, a discriminator is added to the shared encoder and adversarial learning is applied 
  to make sure the shared encoder can learn domain invariant knowledge.
  We conducted experiments on multiple data sets and get consistent significant improvements. We also verified via experiments that the shared encoder, the domain specific private encoder-decoder and the discriminator all make contribution to the performance improvements.

\section*{Acknowledgements}
We thank the three anonymous reviewers for their
comments, Jinchao Zhang, Wen Zhang for suggestions. This work was supported by the National Natural Science Foundation of China (NSFC) under the project NO.61876174, NO.61662077 and NO.61472428.


\bibliography{naaclhlt2019}
\bibliographystyle{acl_natbib}

\end{document}